%% file: example_paper.tex
\documentclass{article}
\input{math_commands.tex}
\usepackage{microtype}
\usepackage{graphicx}
\usepackage{subcaption}
\usepackage{booktabs} 
\usepackage{subfiles}
\usepackage{hyperref}
\usepackage{pifont}
\usepackage{tikz}
\usepackage{xcolor}
\usepackage[hang,flushmargin]{footmisc}
\DeclareMathOperator{\arcsinh}{arcsinh}

\usepackage[accepted]{icml2023}

\usepackage{amsmath}
\usepackage{amssymb}
\usepackage{mathtools}
\usepackage{amsthm}

\usepackage[capitalize,noabbrev]{cleveref}

\theoremstyle{plain}
\newtheorem{theorem}{Theorem}[section]

\newtheorem{lemma}[theorem]{Lemma}
\newtheorem{corollary}[theorem]{Corollary}
\theoremstyle{definition}

\theoremstyle{remark}

\usepackage[textsize=tiny]{todonotes}

\icmltitlerunning{MonoFlow: Rethinking Divergence GANs via the Perspective of Wasserstein Gradient Flows}

\begin{document}

\twocolumn[
\icmltitle{MonoFlow: Rethinking Divergence GANs via the \\Perspective of Wasserstein Gradient Flows}



\icmlsetsymbol{equal}{*}

\begin{icmlauthorlist}
\icmlauthor{Mingxuan Yi}{bristol}
\icmlauthor{Zhanxing Zhu}{cpnl,Peking}
\icmlauthor{Song Liu}{bristol}

\end{icmlauthorlist}

\icmlaffiliation{bristol}{University of Bristol.}
\icmlaffiliation{cpnl}{Changping National Laboratory, China.}
\icmlaffiliation{Peking}{Peking University}

\icmlcorrespondingauthor{Mingxuan Yi}{mingxuan.yi@bristol.ac.uk}

\icmlkeywords{Machine Learning, ICML}

\vskip 0.3in
]



\printAffiliationsAndNotice{}  

\begin{abstract}
The conventional understanding of adversarial training in generative adversarial networks (GANs) is that the discriminator is trained to estimate a divergence, and the generator learns to minimize this divergence. We argue that despite the fact that many variants of GANs were developed following this paradigm, the current theoretical understanding of GANs and their practical algorithms are inconsistent. In this paper, we leverage Wasserstein gradient flows which characterize the evolution of particles in the sample space, to gain theoretical insights and algorithmic inspiration for GANs. We introduce a unified generative modeling framework – MonoFlow: the particle evolution is rescaled via a monotonically increasing mapping of the log density ratio. Under our framework, adversarial training can be viewed as a procedure first obtaining MonoFlow's vector field via training the discriminator and the generator learns to draw the particle flow defined by the corresponding vector field.
We also reveal the fundamental difference between variational divergence minimization and adversarial training. This analysis helps us to identify what types of generator loss functions can lead to the successful training of GANs and suggest that GANs may have more loss designs beyond the literature (e.g., non-saturated loss), as long as they realize MonoFlow. Consistent empirical studies are included to validate the effectiveness of our framework.
\end{abstract}

\subfile{1.Introduction}
\subfile{2.Background}

\subfile{3.Method}

\subfile{4.Results}

\subfile{5.Generator_loss}
\subfile{6.RW_con}



\bibliography{example_paper}
\bibliographystyle{icml2023}

\newpage
\appendix
\onecolumn

\subfile{7.Appendix}

\end{document}

%% file: math_commands.tex
\usepackage{amsmath,amsfonts,bm}

\def\eqref#1{equation~\ref{#1}}

\def\1{\bm{1}}

\def\rvtheta{{\mathbf{\theta}}}

\def\rvw{{\mathbf{w}}}
\def\rvx{{\mathbf{x}}}

\def\rvz{{\mathbf{z}}}
\def\gradx{{\nabla_{\rvx}}}

\DeclareMathAlphabet{\mathsfit}{\encodingdefault}{\sfdefault}{m}{sl}
\SetMathAlphabet{\mathsfit}{bold}{\encodingdefault}{\sfdefault}{bx}{n}

\def\gF{{\mathcal{F}}}

\newcommand{\sigmoid}{\sigma}

\newcommand{\xmark}{\ding{55}}%

\def\pd{p_{\rm{data}}}

\def\jocobgenerator{\frac{\partial g_{\rvtheta_t}(\rvz)}{\partial \rvtheta_t} }
\def\jocobgeneratorz{\frac{\partial g_{\rvtheta_t}(\rvz')}{\partial \rvtheta_t} }
\def\Epz{\mathbb{E}_{\rvz\sim p_\rvz}}

%% file: 1.Introduction.tex
\section{Introduction}
\label{sec:intro}
Generative adversarial nets (GANs) \citep{goodfellow2014generative, jabbar2021survey} are a powerful generative modeling framework that has gained tremendous attention in recent years. GANs have achieved significant successes in applications, especially in high-dimensional image processing such as high-fidelity image generation \citep{brock2018large, karras2019style}, super-resolution \citep{ledig2017photo} and domain adaption \citep{zhang2017stackgan}. 

\begin{figure}[!htb]
\vspace{-2.0mm}
     \centering
     \begin{subfigure}{0.20\textwidth}
         \centering
         \includegraphics[width=\textwidth]{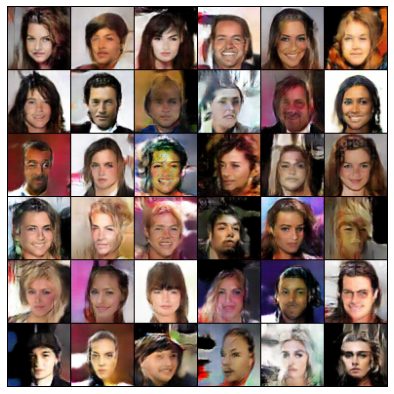}
         \vspace{-6mm}
        
     \end{subfigure}
    \hspace{1.0mm}
     \begin{subfigure}{0.20\textwidth}
         \centering
         \includegraphics[width=\textwidth]{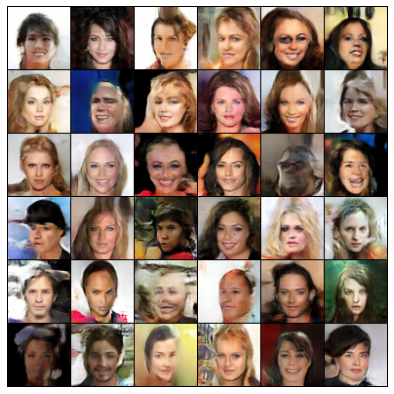}
         \vspace{-6mm}

     \end{subfigure}
\caption{Generated Celeb-A faces \citep{liu2015deep} with the logit loss and the $\arcsinh$ loss.}
\label{faces}
\end{figure}

In the GAN framework, a discriminator $d$ and a generator $g$ play a minmax game. The discriminator is trained to distinguish real and fake samples and the generator is trained to generate fake samples to fool the discriminator. The equilibrium of the vanilla GAN is defined by\footnote{We use a slightly different notation: $d(\rvx)$ is the logit output of the classifier and $\sigmoid(\cdot)$ is the Sigmoid activation.}
\begin{equation}
\begin{split}
\min_g \max_d V(g, d) = &\mathbb{E}_{\rvx\sim \pd}\big\{\log\sigmoid [d(\rvx)]\big\}+ \\ &\quad \mathbb{E}_{\rvz\sim p_\rvz}\left\{\log \big(1-\sigmoid [d(g(\rvz))]\big) \right\} \label{adv_gam}
\end{split}
\end{equation}
The elementary optimization approach to solve the minmax game is adversarial training.
Previous perspectives explained it as first estimating Jensen-Shannon divergence and the generator learns to minimize this divergence. Several variants of GANs have been developed based on this point of view for other probability divergences, e.g., $\chi^2$ divergence \citep{mao2017least}, Kullback-Leibler (KL) divergence \citep{arbel2020generalized} and general $f$-divergences \citep{nowozin2016f, uehara2016generative}, while others are developed with Integral Probability Metrics \citep{arjovsky2017wasserstein, dziugaite2015training, mroueh2017sobolev}. However, we emphasize that the traditional perspective on GANs is inconsistent and we present three non-negligible facts which are commonly associated with adversarial training, making it markedly different from the standard variational divergence minimization (VDM) problem: 
\begin{enumerate}
    \item The estimated divergence is computed from the discriminator $d(\rvx)$. $d(\rvx)$ is trained using samples $\rvx$ only such that it cannot capture the variability of the generator's distribution $p_g$ \citep{metz2016unrolled, franceschi2022neural}. However, the optimal discriminator in the adversarial game by \citet{goodfellow2014generative} requires $p_g$ to be a functional variable such that the dependency between the optimal discriminator and $p_g$ exists, i.e., the discriminator is a function $d(\rvx, g)$ taking as input generator's parameter as well.  
    \item The generator minimizes a divergence with a missing term, e.g., the vanilla GAN only minimizes the second term of the Jensen-Shannon divergence $-\mathbb{E}_{\rvz\sim p_\rvz}\left\{-\log \big(1-\sigmoid [d(g(\rvz))]\big) \right\}$ which is, however, a KL divergence up to a constant, see Eq.~(5) of \citet{goodfellow2014generative}.
    \item Practical algorithms are inconsistent with the theory, a heuristic trick ``non-saturated loss" is commonly adopted to mitigate the gradient vanishing problem, but it still lacks a rigorous mathematical understanding. For example, the generator loss of the non-saturated GAN is $-\mathbb{E}_{\rvz\sim p_\rvz}\big\{\log \sigmoid [d(g(\rvz))]\big\}$. We can even modify the generator loss to the logit loss $-\mathbb{E}_{\rvz\sim p_\rvz}\big\{  d(g(\rvz))\big\}$ or the $\arcsinh$ loss $-\Epz \left\{\arcsinh \big(d(g(\rvz))\big) \right\}$, the generator still learns the data distribution, as shown in Figure \ref{faces}.
\end{enumerate}
All of the above generator losses satisfy 
\begin{equation}
    -\mathbb{E}_{\rvz\sim p_\rvz}\big\{h [d(g(\rvz))] \big\},
\end{equation}
where $h\colon \mathbb{R} \to \mathbb{R}$ is a strictly monotonically increasing function with $h'(\cdot)>0$. It is known the logit output $d(\rvx)$ of a binary classifier in Eq.~(\ref{adv_gam}) is the logarithm density ratio estimator between two distributions \citep{Qin1998,sugiyama2012density}. To gain a deeper understanding of divergence GANs, we study the Wasserstein gradient flow of the KL divergence which characterizes a Euclidean particle flow ordinary differential equation (ODE). This ODE is also known as the ``probability flow ODE" \citep{song2020score} of Langevin dynamics, with its vector field defined by the gradient of the log density ratio. Based on this ODE, we propose the \textbf{MonoFlow} framework -- transforming the log density ratio by a strictly increasing mapping such that the vector field of the ODE is rescaled along the same direction. Consequently, learning to simulate MonoFlow is identical to training divergence GANs. 
All variants of divergence GANs are a subclass of our framework. We reveal that the discriminator loss and generator loss do not need to follow the same objective which is contradictory to the adversarial game \citep{goodfellow2014generative}. The discriminator maximizes an objective to obtain a bijection of the log density ratio. Then the generator loss can be any strictly increasing mapping of this learned log ratio. Our contributions can be summarized as follows:  
\begin{itemize}
    \item A novel generative modeling framework has been developed, which unifies divergence GANs and provides a new understanding of their training dynamics. This framework not only provides a new theoretical perspective but also ensures practical consistency.
    \item We reveal the fundamental difference between VDM and adversarial training, which indicates that the previous analysis of GANs based on the perspective of VDM might not provide benefits, and instead we should treat GANs as a particle flow method similar to diffusion models \citep{ho2020denoising,song2020score}.
    \item An analysis of what types of generator losses can practically lead to the success of training GAN. Our framework explains why and how non-saturated loss works. 
    \item An algorithmic inspiration where GANs may have more variants of generative losses than we already know. 
\end{itemize}



%% file: 2.Background.tex
\section{Wasserstein Gradient Flows} \label{wgf_section}
In this section, we review the definition of gradient flows in Wasserstein space $(\mathcal{P}(\mathbb{R}^n), W_2)$, the space of Borel probability measures $\mathcal{P}(\mathbb{R}^n)$ defined on $\mathbb{R}^n$ with finite second moments and equipped with the Wasserstein-2 metric. An absolutely continuous curve of probability measures $\{q_t\}_{t\geq 0}\in \mathcal{P}(\mathbb{R}^n)$ is a Wasserstein gradient flow if it satisfies the following continuity equation \citep{ambrosio2005gradient},
\begin{equation}
    \frac{\partial {q_t}}{\partial t} = \text{div}\big(q_t \nabla_{W_2} {\mathcal{F}(q_t)}\big), \label{wgf}
\end{equation}
where $\nabla_{W_2} {\mathcal{F}(q_t)}$ is called the Wasserstein gradient of the functional $\mathcal{F} \colon \mathcal{P}(\mathbb{R}^n) \to \mathbb{R}$. 
\begin{figure*}[h]
\vspace{-0.5mm}
\begin{tikzpicture} 
  \node[right] at (-3.6, 1.2) {\normalsize Wasserstein space:};
  \node[right] at (5.5, 1.2) {\normalsize Euclidean space:};
  \draw[-stealth, domain=-3:3,smooth,variable=\x, cyan!80!black, ultra thick] plot ({\x},{-0.15*\x*\x });
  \draw [fill, cyan!80!black] (-3,-9*0.15) circle [radius=0.1] ;
  \node[right] at (-0.0,-2.25*0.15+0.3*1.5*1.5) {\normalsize $-\nabla_{W_2} {\mathcal{F}(q_t)}$};
  
  \node[right] at (3,-9*0.15) {\normalsize $p$};
  \node[left] at (-3,-9*0.15) {\normalsize $q_0$};
  \node[above] at (-1.5,-2.25*0.15) {\normalsize $q_t$};
  \filldraw[color=blue!60, fill=blue!5, very thick](6.0,0.5) circle (0.4);
  \filldraw[color=blue!60, fill=blue!5, very thick](10.05,0.2-9*0.15+0.1) circle (0.4);
  \draw [fill, black] (-1.5,-2.25*0.15) circle [radius=0.1] ;
  \draw[-stealth, very thick] (-1.5,-2.25*0.15)--(-0.0,-2.25*0.15+0.3*1.5*1.5);

  \draw [dotted, very thick, red] (-3,-9*0.15) to (3,-9*0.15);
  \draw[thick, red] (-3, -9*0.15+0.2) -- ++ (0.0,-0.4);
  \draw[thick, red] (3, -9*0.15+0.2) -- ++ (0.0,-0.4);
  \node[below, red] at (0, -9*0.15) {\normalsize $W_2(q_0, p)$};

\draw [fill, black] (6,  .7) circle [radius=0.05] ;
\draw [fill, black] (5.9,0.4) circle [radius=0.05] ;
\draw [fill, black] (6.2,0.4) circle [radius=0.05] ;
 \node[above] at (7.0,0.4) {$\rvx_t \sim q_t$};
 \draw[-stealth, very thick]  (6.2,0.4)--(7.5, -0.3);
 \node[right] at (7.5, -0.1) {\normalsize $v_t=-\gradx \frac{\delta \mathcal{F}}{\delta{q_t}}(\rvx)\Big\vert_{\rvx=\rvx_t}$};
 \draw [fill, black] (10,  .5-9*0.15) circle [radius=0.05] ;
\draw [fill, black] (9.9,0.2-9*0.15) circle [radius=0.05] ;
\draw [fill, black] (10.2,0.2-9*0.15) circle [radius=0.05] ;
\node[right] at (10.5,0.2-9*0.15) {\normalsize $p$};
\end{tikzpicture}
\vspace{-2.5mm}
\caption{The illustration of a Wasserstein gradient flow and its particle evolution. In Wasserstein space, the blue curve is a gradient flow and the red dotted line is a geodesic. $q_t$ evolves along a curve whose tangent vector is given by $-\nabla_{W_2} {\mathcal{F}(q_t)}$ such that the functional is always decreasing with time. Correspondingly, particles evolve in Euclidean space towards the target measure $p$ with the vector field $-\gradx\frac{\delta \mathcal{F}}{\delta{q_t}}(\rvx)$. Note that directly minimizing the Wasserstein-2 metric $W_2(q_t, p)$ instead yields a path $\{q_t\}_{t\geq0}$ along the geodesic connecting $q_0$ and $p$ over Wasserstein space.}\vspace{-3mm}
\label{wgf_fig}
\end{figure*}

The Wasserstein gradient is given as $\nabla_{\rvx}(\delta \mathcal{F}/{\delta{q_t}})$, i.e.  the Euclidean gradient of the functional's first variation $\delta \mathcal{F}/{\delta{q_t}}$. Specifically, for the KL divergence $\mathcal{F}(q_t)=\int \log (q_t/p) \mathrm{d}q_t$, 
where $p$ is a fixed target probability measure, we have $\delta \mathcal{F}/{\delta{q_t}}= \log q_t - \log p + 1$. Hence, the Wasserstein gradient flow of the KL divergence reads the Fokker-Planck equation \citep{risken1996fokker},
\begin{equation}
    \frac{\partial {q_t}}{\partial t} = \text{div}\big( q_t (\gradx \log q_t-\gradx \log p) \big). \label{fk Eq.}
\end{equation} 
If we denote the Euclidean path of random variables as $\{\rvx_t\}_{t \geq 0} \in \mathbb{R}^n$ with the initial condition $\rvx_0 \sim q_0$, we can define an ordinary differential equation (ODE) to describe the evolution of particles in $\mathbb{R}^n$,  \begin{equation}
    \mathrm{d}\rvx_t = \big( \gradx \log p(\rvx_t) - \gradx \log q_t(\rvx_t) \big)\mathrm{d}t:=v_t(\rvx_t)\mathrm{d}t, \label{kl flow}
\end{equation}
where the vector field $v_t$ of these particles is the negative  Euclidean gradient of the functional's first variation.
 As shown in Figure \ref{wgf_fig}, Wasserstein gradient flows establish a connection between the probability evolution in Wasserstein space and its associated particle evolution in Euclidean space.  

Applying It\^{o} integral to Langevin dynamics $\mathrm{d}\rvx_t = \gradx \log p(\rvx_t)\mathrm{d}t + \sqrt{2} \mathrm{d}\rvw_t$ where $\mathrm{d}\rvw_t$ is a Wiener process, we obtain the same Fokker-Planck equation in Eq.~(\ref{fk Eq.}). This indicates that the deterministic particle evolution by the ODE can be approximated via a stochastic differential equation (SDE). Langevin dynamics admits the same marginal probability measure $q_t$ as Eq.~(\ref{kl flow}), this relation of SDE and its corresponding ODE was also studied in score-based diffusion models \citep{song2020score}. Langevin dynamics was first interpreted as the Wasserstein gradient flow of the KL divergence by \citet{jordan1998variational, otto2001geometry}. It plays an important role in generative modeling as a sampling scheme. In order to transform noises into the target data distribution by Langevin dynamics, an essential step is to fit the data distribution using energy-based models \citep{song2021train} or to directly estimate its scores with score-matching techniques \citep{hyvarinen2005estimation, vincent2011connection, song2019generative}.


%% file: 3.Method.tex
\section{MonoFlow: A Unified Generative Modeling Framework}
This section presents our main contribution that connects gradient flows and divergence GANs. We first introduce \textbf{MonoFlow} where the ODE evolution is rescaled via a monotonically increasing function. Consequently, learning to simulate and draw the rescaled particle flow recovers the bi-level optimization dynamics of training divergence GANs. This gives us a novel understanding of the hidden mechanism of adversarial training. 
\subsection{MonoFlow}
We consider the ODE in Eq.~(\ref{kl flow}) with a fixed target measure $p$, e.g., a data distribution in a generative modeling scenario. Assume that we have a time-dependent log density ratio function as $\log r_t(\rvx) = \log \big[p(\rvx)/q_t(\rvx)\big]$, the ODE can be rewritten as 
\begin{equation}
\mathrm{d}\rvx_t = \nabla_{\rvx} \log r_t(\rvx_t) \mathrm{d}t, \quad \rvx_t \sim q_t. \label{log ratio flow}
\end{equation}
This is a gradient flow in Euclidean space where its vector field is the gradient of the log density ratio. With a strictly monotonically increasing and differentiable mapping $h\colon \mathbb{R} \to \mathbb{R}$, we can define another ODE: 
\begin{equation} 
\begin{split}
     \mathrm{d}\rvx_t &= \nabla_{\rvx} h\big(\log r_t(\rvx_t)\big) \mathrm{d}t \\
     &= h'\big(\log r_t(\rvx_t)\big)\nabla_{\rvx} \log r_t(\rvx_t) \mathrm{d}t, \quad \rvx_t \sim q_t \label{mono gf}
     \end{split}
\end{equation}
By transforming the time-dependent log density ratio under the mapping $h$, its first-order derivative rescales the vector field of the original particle flows defined in Eq.~(\ref{log ratio flow}). We call Eq.~(\ref{mono gf}) as \textbf{MonoFlow}.
\begin{table*}[t]
\caption{Different types of divergence GANs. $f$ is a convex function and $\tilde{f}$  is the convex conjugate by $\tilde{f}(d) = \sup_{r\in\text{dom}f}\{rd - f(r)\}$.}
\vspace{-0.5mm}
\label{GAN_variants}
\begin{center}
\begin{tabular}{lcccc}
\multicolumn{1}{c}{}   &$\phi(d)$ &$\psi(d)$   &$d^*(\rvx)$ & $h_{\mathcal T}(d)$\\
\toprule 
Vanilla GAN        &$\log \sigmoid(d)$  &$\log (1-\sigmoid(d))$  &$\log r(\rvx)$               &$-\log (1-\sigmoid(d))$ \\
Non-saturated GAN   &$\log \sigmoid(d)$  &$\log (1-\sigmoid(d))$  &$\log r(\rvx)$               &$\log \sigmoid(d)$       \\
$f$-GAN            &$d$                 &$-\tilde{f}(d)$          &$f'(r(\rvx))$               & $d$        \\
$b$-GAN            &$f'(d)$             &$f(d)-df'(d)$           &$r(\rvx)$                    & $df'(d)-f(d)$  \\
Least-square GAN   &$-(d-1)^2$           &$-d^2$                   &$\frac{r(\rvx)}{1+r(\rvx)}$  & $-(d-1)^2$     \\
Generalized EBM (KL) &$-(d + \lambda)$ & $-\exp(-d-\lambda)$ &$-\log r(\rvx) - \lambda$ & $\exp(-d-\lambda)$\\
\bottomrule
\end{tabular}
\end{center}

\vspace{-3mm}
\end{table*}
MonoFlow defines a different family of vector fields $\{v_t\}_{t\geq 0}$ for the particle evolution where $v_t(\rvx_t)= h'\big(\log r_t(\rvx_t)\big)\nabla_{\rvx} \log r_t(\rvx_t)$. Conversely, the vector fields $\{v_t\}_{t\geq 0}$ also determine an absolutely continuous curve $\{q_t\}_{t\geq 0}$ in Wasserstein space by the continuity equation (see Theorem 4.6 of \citet{ambrosio2005gradient}), 
\begin{equation} 
\frac{\partial q_t}{\partial t} = -\text{div} (q_t v_t),
\end{equation}
under mild regularity conditions.
Hence the probability evolution of MonoFlow is described by 
\begin{equation}
    \frac{\partial q_t}{\partial t} = \text{div} \big(D_t \gradx q_t \big) - \text{div} \big( \zeta^{-1}_t q_t \gradx \log p \big), \label{cd equation}
\end{equation}
where $D_t = \zeta^{-1}_t =h'(\log r_t)$. Eq.~(\ref{cd equation}) is a special case of convection-diffusion equations where $D_t$ is called the diffusion coefficient and $\zeta^{-1}_t$ is called mobility. MonoFlow defines a positive diffusion coefficient. This has a physical interpretation that particles diffuse to spread probability mass over the target measure other than concentrate. Next, we study the properties of MonoFlow. Proofs are provided in Appendix \ref{appA}.  
\begin{theorem}
\label{theorem1}
If  $h'(\cdot) > 0$, the dissipation rate ${\partial \mathcal{F}(q_t)}/{\partial t}$ for the KL divergence $\mathcal{F}(q_t)=\int \log \big({q_t}/{p}) dq_t$ satisfies
\begin{equation}
\frac{\partial \mathcal{F}(q_t)}{\partial t} \leq 0,
\end{equation}
the equality is achieved if and only if $q_t = p$ and the marginal probability $q_t$ of MonoFlow evolves to $p$ as $t \to \infty$.
\end{theorem} 
Theorem \ref{theorem1} shows that MonoFlow does not disturb the stationary measure of Eq.~(\ref{fk Eq.}). The negative dissipation rate ensures that the curve $\{q_t\}_{t \geq0} $ of MonoFlow always decreases the KL divergence with time. It is obvious that the marginal probability $q_t$ finally evolves to the target $p$ with time since the KL divergence converges to zero if $t\to \infty$, by the monotone convergence theorem. Note that we do not assume the target probability measure $p$ is log-concave, the rate of convergence is not studied in this paper. 


MonoFlow is obtained by transforming the log density ratio which arises from the Wasserstein gradient flow of the KL divergence. We can also formulate different deterministic particle evolution by considering Wasserstein gradient flows of general $f$-divergences, 
\begin{equation}
 \mathcal{D}_{f}(p||q_t)=  \int  f\left(r_t \right)dq_t,\quad r_t=\frac{p}{q_t}
\end{equation}
where $f\colon \mathbb{R}^+ \to \mathbb{R}$ is a twice differentiable convex function with $f(1)=0$.
\begin{theorem} 
\label{theorem2}
The Wasserstein gradient flow of an $f$-divergence characterizes the evolution of particles in $\mathbb{R}^n$ by
\begin{equation}
\mathrm{d}\rvx_t = r_t(\rvx_t)^2f''\big(r_t(\rvx_t)\big)\nabla_{\rvx} \log r_t(\rvx_t) \mathrm{d}t, \quad \rvx_t \sim q_t.
\end{equation}
\end{theorem} 
A similar result can also be derived with the reversed $f$-divergences $ \mathcal{D}_{f}(q_t||p)$ used by \citet{johnson2018composite, gao2019deep, ansari2020refining}. Theorem \ref{theorem2} shows that the particle evolution of the Wasserstein gradient flow of $f$-divergences is a special instance of MonoFlow if a stronger condition than convexity is holding, i.e., $f''(\cdot) >0 $ which implies $f$ is a strictly convex function. The rescaling factor is given by $h'(\log r) = r^2 f''(r) > 0$, this indicates once a curve $\{q_t\}_{t\geq 0 }$ evolves with the time $t$ in Wasserstein space to decrease an $f$-divergence whose $f''(\cdot)>0$, it simultaneously decreases the KL divergence as well since the dissipation rate of MonoFlow is negative. 

Furthermore, a corollary of Theorem \ref{theorem2} is that MonoFlow implicitly defines Wasserstein gradient flows of $f$-divergences via the increasing function $h$ without specifying any strictly convex functions $f$.
\begin{corollary}
For any continuously differentiable $h\colon \mathbb{R} \to \mathbb{R}$ with $h'(\cdot) > 0$ , there exists a strictly convex and twice differentiable function $f\colon \mathbb{R}^+ \to \mathbb{R}$ with $f(1)=0$ satisfying
\begin{equation}   
h(\log r) = rf'(r) - f(r),
\end{equation} 
MonoFlow associated with this increasing function $h$ is the Wasserstein gradient flow of the functional $\mathcal{F}(q) = \mathcal{D}_{f}(p||q)$.
\end{corollary} \label{coro}


\subsection{Practical Approximations of Density Ratios}
We first discretize the ODE in Eq.~(\ref{mono gf}) by the forward Euler method such that we obtain standard gradient ascent iterations with step size $\alpha$ and the index of the discretized time step $k$ \footnote{For the sake of simplicity, we briefly replace $t_k$ by its index $k$, though it is not rigorous.}:
\begin{equation}
    \rvx_{k+1} =\rvx_{k} + \alpha\gradx h\big(\log r_k(\rvx_k)\big) , \quad t_{k+1} = t_k + \alpha.
\end{equation}
Therefore, we can sample initial particles $\rvx_0 \sim q_0$ and perform gradient ascent iterations by estimating the density ratio $r_k(\rvx) = {p(\rvx)}/{q_k(\rvx)}$ using samples from $q_k$ and $p$.
In order to enable a practical algorithm to obtain the time-dependent density ratio, we introduce a general framework that solves the following optimization problem,
\begin{equation}
\max_{d \in \mathcal{H}} \big\{\mathbb{E}_{\rvx\sim p}\left[\phi\big(d(\rvx)\big)\right] + \mathbb{E}_{\rvx\sim q_k}\left[\psi \big( d(\rvx)\big)\right]\big\}, \label{general dr} 
\end{equation}
where $d\colon \mathbb{R}^n \to \mathbb{R}$ is a discriminator and $\mathcal{H}$ is a class of all measurable functions. $\phi$ and $\psi$ are differentiable scalar functions upon design later. Similar to \citep{moustakides2019training}, we show that if $\phi$ and $\psi$ satisfy conditions in Lemma \ref{lemm3}, the optimal $d^\ast$ is a bijection of the density ratio between $p$ and $q_k$.
\begin{lemma}
\label{lemm3}
Define $\mathcal{T}(d(\rvx))  :=-\frac{\psi'(d(\rvx))}{\phi' ( d(\rvx)) } $. If $\phi$ and $\psi$ satisfy either of
\begin{enumerate}
    \item $\phi$ is concave, $\psi$ is strictly concave and the mapping $\mathcal{T}$ is a bijection.
    \item $\phi'(\cdot) >0 $ and the mapping $\mathcal{T}$ is strictly increasing (also a bijection).
\end{enumerate}
Solving Eq.~(\ref{general dr}), the optimal $d^*$ satisfies 
\begin{equation}
d^*(\rvx) = \mathcal{T}^{-1}(r(\rvx)),\quad r(\rvx) = \frac{p(\rvx)}{q_k(\rvx)},
\end{equation}
\end{lemma}

\textbf{Remark:} Note that two-sample density ratio estimations discard the density information from $q_k$. The functions $d(\rvx)$, $r(\rvx)$ only depend on $\rvx$ and they cannot capture the variability of $q_k$. 

To this end, we can train $d$ to solve the optimization problem in Eq.~(\ref{general dr}) and the density ratio is approximated by $\mathcal{T}(d(\rvx))$. For example, in a binary classification problem where $\phi(d) = \log \sigmoid(d)$ and $\psi(d) = \log (1-\sigmoid(d))$, we have $d^*(\rvx) = \log r(\rvx)$ where its post-Sigmoid output $\sigma(d^*(\rvx)) = {r(\rvx)}/{\big(1+r(\rvx)\big)}= {p(\rvx)} /\big({p(\rvx)+q_k(\rvx)}\big)$ is aligned with the Proposition 1 of \citet{goodfellow2014generative}.  Other types of density ratio estimation can be found in Table \ref{GAN_variants} as they have been already used in GAN variants where $q_k$ refers to the generator's distribution $p_g$. Specifically, $f$-GAN \citep{nowozin2016f}, Least-square GAN\citep{mao2017least}, Generalized EBM \cite{arbel2020generalized} satisfy the condition 1 and $b$-GAN \citep{uehara2016generative} satisfies the condition 2 in Lemma \ref{lemm3}. 

 In practice, since the change of $\rvx_k$ is sufficiently small at every step $k$, we can use a single discriminator $d(\rvx)$ and perform a few gradient updates to solve Eq.~(\ref{general dr}) per iteration to approximate the time-dependent density ratio $r_k(\rvx)$, which is identical to the GAN training.

\subsection{Parameterization of the Discretized MonoFlow}
The previous method directly pushes particles in the Euclidean space towards the target measure. We can use a neural network generator to mimic the distribution of these particles, i.e., train the generator to learn to draw samples. 

We parameterize particles with a neural network generator $g_{\rvtheta_k}$ that takes as input random noises $\rvz \sim p_\rvz$ and output particles $\rvx_k = g_{\rvtheta_k}(\rvz)$, we next move particles along the vector field of MonoFlow, 
\begin{equation}
    \rvx_{k+1} = g_{\rvtheta_k}(\rvz) + \alpha \gradx h\big(\log r_k(g_{\rvtheta_k}(\rvz) ) \big)
\end{equation}
Similar to \citep{wang2016learning}, in order to encourage the generator to draw particles more similar to $\rvx_{k+1}$, we use one-step gradient descent to approximately solve $\min_{\rvtheta} \mathbb{E}_{\rvz \sim p_\rvz} ||g_{\rvtheta}(\rvz) - \rvx_{k+1}||^2$, such that the generator's parameter is updated with learning rate $\beta$ via
\begin{equation}
\theta_{k+1} 
= \theta_k + \beta  \nabla_{\rvtheta}\Epz \big[ h\big(\log r_k(g_{\rvtheta_k}(\rvz) ) \big)\big].\label{armo}
\end{equation}
In the continuous-time evolution, the associated infinitesimal change of the generator's parameter can be written as 
\begin{equation}
\frac{\mathrm{d}{\rvtheta_t}}{\mathrm{d}t} = \int \jocobgenerator \gradx h\big(\log r_t(\rvx_t) \big) p_\rvz(\rvz) \mathrm{d}\rvz \label{dtheta},
\end{equation}
where $\jocobgenerator$ is the Jacobian of the neural network generator. Consequently, if particles are generated via $\rvx_t = g_{\rvtheta_t}(\rvz)$, we have $\mathrm{d}\rvx_t = \jocobgenerator \mathrm{d}{\rvtheta_t}$ by the chain rule, replace $\mathrm{d}{\rvtheta_t}$ with Eq.~(\ref{dtheta}), we obtain
\begin{equation}
\mathrm{d}\rvx_t = \mathbb{E}_{\rvz' \sim p_\rvz} \left[K^t_{g}(\rvz, \rvz') \gradx h\big(\log r_t(\rvx_t) ) \right]\mathrm{d}t \label{ntkflow}
\end{equation}
where $K^t_{g}(\rvz, \rvz') = \langle \jocobgenerator, \jocobgeneratorz \rangle$ is the neural tangent kernel (NTK) \citep{jacot2018neural} defined by the generator. Eq.~(\ref{ntkflow}) realizes \textit{Stein Variational Gradient Descent} \citep{liu2016stein, franceschi2022neural} if $h$ is an identity mapping. 

\subsection{A Unified Formulation of Divergence GANs}
Based on the above derivation, we propose a general formulation for divergence GANs. We clarify that GANs can be treated with different objective functions for training discriminators and generators. All of these variants are algorithmic instantiations of the parameterized MonoFlow. The unified framework is summarized as: given a discriminator $d$ and a generator $g$, the discriminator $d$ learns to maximize
\begin{equation}
\mathbb{E}_{\rvx\sim \pd}\left[\phi\big(d(\rvx)\big)\right] + \mathbb{E}_{\rvz\sim p_\rvz}\left[\psi \big( d(g(\rvz))\big)\right], \label{vec}
\end{equation}
where $\pd$ refers to the data distribution. Next, we train the generator $g$ to minimize
\begin{equation}
-\Epz \left[h_\mathcal{T}\big(d(g(\rvz))\big)\right].\label{amor}
\end{equation}
where $h_\mathcal{T} (d) = h\big(\log (\mathcal{T}(d))\big)$ and $h$ can be any strictly increasing function with $h'(\cdot)>0$. 
We summarize some typical GAN variants in Table \ref{GAN_variants}. We view adversarial training as maximizing Eq.~(\ref{vec}) to obtain the density ratio estimator which suggests the vector field for MonoFlow and minimizing Eq.~(\ref{amor}) as learning to parameterize MonoFlow corresponding to Eq.~(\ref{armo}). 

%% file: 4.Results.tex
\section{Understanding Adversarial Training via MonoFlow}
The dominating understanding of adversarial training over GANs is that the generator learns to minimize the divergence estimated from the discriminator. However, as pointed out in Section~\ref{sec:intro}, the theoretical explanation of GANs and the practical algorithms are inconsistent. In this section, through the lens of MonoFlow, we will explain why this inconsistency does not prevent divergence GANs from achieving decent results and how it differs from a variational divergence minimization (VDM) problem. 
\subsection{Why the Adversarial Game Works?}
In an adversarial game, the discriminator is trained to maximize the lower bound of $f$-divergences. This lower bound can be derived via the dual representation of $f$-divergences \citep{nguyen2010estimating} between $\pd$ and $p_g$, 
\begin{equation}
\begin{split}
&\mathcal{D}_{f}(\pd||p_g) \\
&\quad \quad\quad = \max_{d\in\mathcal{H}} \big\{ \underbrace{\mathbb{E}_{\rvx\sim \pd}\big[d(\rvx)\big] - \mathbb{E}_{\rvx\sim p_g}\big[\tilde{f} \big( d(\rvx)\big)\big]}_{\text{lower bound}} \big\}, \label{var_f}
\end{split}
\end{equation}
where $r(\rvx)={\pd(\rvx)}/{p_g(\rvx)}$ and $\tilde{f}(d) = \sup_{r\in\text{dom}f}\{rd - f(r)\}$ is the convex conjugate of $f(r)$. Note that for binary classification problems where we design specific $\phi$ and $\psi$, the corresponding optimization problem in Eq.~(\ref{vec}) can be translated into an equivalent formulation as the above dual representation \citep{nowozin2016f}. Since the first term of the lower bound in Eq.~(\ref{var_f}) is irrelevant to $p_g$, the generator actually only learns to minimize the second term (vanilla loss), 
\begin{equation}
 \min_g - \mathbb{E}_{\rvx\sim p_g}\big[\tilde{f} \big( d(\rvx)\big)\big]
\end{equation}
Meanwhile, the generator can also alternatively minimize the heuristic non-saturated loss $ - \mathbb{E}_{\rvx\sim p_g}\big[d(\rvx)\big]$, which has been proven to work well in practice \citep{goodfellow2014generative, nowozin2016f}. By the Fenchel duality, the optimal $d^*$ is given by 
\begin{equation}
    d^*=f'(r) \label{eq:optimal_d}
\end{equation}
with the equality $\tilde{f}(d^*)= rf'(r) - f(r)$. 
Fortunately, it can be simply verified that $f'(r)$ and $rf'(r)-f(r)$ are both strictly increasing functions of the density ratio (as well as the log density ratio) with positive derivatives if $f''(\cdot)>0$ which implies strict convexity of $f$. Hence, adversarial training with the vanilla loss and the non-saturated loss both fall into the framework of MonoFlow which has theoretical guarantees.
\begin{table*}[h]
 \vspace{-4mm}
\caption{Comparisons on three density ratio models using different $f$ and $h$: ``\checkmark"\ means the generator learns the data distribution and ``\xmark"\  means it does not work. The evaluation is based on whether or not the parameter of the generator can finally approximate the target $(\mu_0, s_0)$, with visualizations as the complement.  Visualization results are included in Appendix \ref{visu}. Code is available at \url{https://github.com/YiMX/MonoFlow}.}
\vspace{-0mm}
\label{table_1}
\begin{center}
\begin{tabular}{lccccc}
\multicolumn{1}{c}{}   &if $f$ convex &if $h$ increases  &$r(\rvx, \rvtheta)$ & $r(\rvx, \rvtheta_{\text{de}})$ & $r_{\text{GAN}}(\rvx)$\\
\toprule
KL                      &Yes        &Yes         &\checkmark &\checkmark &\checkmark\\
Forward KL              &Yes        &No          &\checkmark &\xmark     &\xmark    \\
Chi-Square              &Yes        &No          &\checkmark &\xmark     &\xmark    \\
Hellinger               &Yes        &No          &\checkmark &\xmark     &\xmark    \\
Jensen-Shannon          &Yes        &No          &\checkmark &\xmark     &\xmark    \\
Exp                     &No         &Yes         &\xmark     &\checkmark &\checkmark    \\
\bottomrule
\end{tabular}
\end{center}
\vspace{-3mm}
\end{table*}


\subsection{Difference between Adversarial Training and Variational Divergence Minimization}
In this part, we show that VDM differs from GAN training because it relies on the dependence between the discriminator (or the bijective density ratio) and the generator's distribution $p_g$. \citet{metz2016unrolled} and \citet{ franceschi2022neural} also noticed that this dependence is discarded during the practical algorithms of GANs. We provide a further discussion of how this issue results in the difference between adversarial training and VDM as elaborated in the following.

The generator of GANs is a black box sampler without defining an explicit density function. However, in a VDM problem, the generator's output $\rvx$ can be reparameterized, e.g., as a Gaussian random variable where $\rvtheta$ are its mean and scale, such that the generator $g_{\rvtheta}$ defines a distribution via an explicit density function $p_g(\rvx;\rvtheta)$. In VDM, we are interested in minimizing an $f$-divergence with respect to $p_g$, 
\begin{equation}
\min_{p_g} \mathcal{D}_{f}(\pd||p_g). \label{mcgrad}
\end{equation}
With the explicit density function $p_g(\rvx;\rvtheta)$, the density ratio $r(\rvx, \rvtheta) = {\pd(\rvx)}/{ p_g(\rvx;\rvtheta)}$ is a function depending on $\rvx$ as well as the generator's parameter $\rvtheta$ to capture the variability of $p_g$. After integrating out $\rvx$, the $f$-divergence can be written as a cost function of $\theta$. 
\begin{equation}
\mathcal{D}_{f}(\pd||p_g) = \mathbb{E}_{\rvx\sim p_g}\big[ f\big(r(\rvx, \rvtheta)\big)\big] =\text{Cost}(\rvtheta)
\end{equation}
Hence, the optimization over the functional space of $p_g$ can be achieved by optimizing the parameter of the generator, 
\begin{equation}
\min_{p_g} \mathcal{D}_{f}(\pd||p_g) \Longleftrightarrow \min_{\rvtheta} \text{Cost}(\rvtheta) .
\end{equation}
Since $f$ is convex, by Jensen's inequality this cost is minimized at zero where $r(\rvx, \theta)$ is a constant for each $\rvx$, meaning $p_g=\pd$ (see details in Appendix \ref{videtail}). 
Similarly, we can rewrite the $f$-divergence under Fenchel-duality as 
\begin{equation}
    \text{Cost}(\theta) = \mathbb{E}_{\rvx\sim \pd}\big[d^*(\rvx, \rvtheta)\big] - \mathbb{E}_{\rvx\sim p_g}\big[\tilde{f} \big( d^*(\rvx, \rvtheta)\big)\big], \label{dual_mcgrad}
\end{equation}

 where $d^*(\rvx, \rvtheta) = f'(r(\rvx, \rvtheta))$. 

The cost function in Eq.~(\ref{dual_mcgrad}) is different from the objective of practical adversarial training in Eq.~(\ref{var_f}) since the first term of Eq.~(\ref{dual_mcgrad}) has a dependence on the generator's parameter $\theta$. This dependence is required in the theoretical adversarial game, see Eq.~(4) of \citet{goodfellow2014generative} as a special case of Eq.~(\ref{dual_mcgrad}).
However, in the practical algorithm, the density ratio estimator $r(\rvx)$ or its bijection $d(\rvx)$ are only functions of the sample $\rvx$.  Plugging $r(\rvx)$ or $d(\rvx)$ into the $f$-divergence to replace $r(\rvx, \rvtheta)$ or $d^*(\rvx, \rvtheta)$, we can recover the approximated $f$-divergences but the approximated divergences can never be viewed as a cost function of $\theta$ anymore. Dropping out this dependence, the generator of GANs only minimizes the second term of the dual form of $f$-divergences or the non-saturated loss heuristically. 
This is the major disconnection between the theory and the practical algorithm over GANs. 


\subsection{Empirical Study of 2D Gaussians} \label{4.3}
 
We show how the dependence between the ratio model and the generator's parameter practically affects matching $p_g$ to $\pd$ on toy data sets. Let the data distribution be a Gaussian $\pd = N(\mu_0, \Sigma_0)$ where $\Sigma_0=s_0^Ts_0$. We start from the simplest form of a generator (reparameterization), $$p_g:\rvx_\theta(\rvz)=g_{\rvtheta}(\rvz)=\mu + s \cdot\rvz, \quad \rvz \sim N(0, I),$$ where $\rvtheta = (\mu, s)$, $\mu$ is the mean and $s$ is the scale matrix. 

By assuming the generator and data distributions are Gaussians, we can define three density ratio models. The first model is $r(\rvx, \rvtheta) = {\pd(\rvx)}/{p_g(\rvx;\rvtheta)}$, where the density ratio function depends on $\rvx$ and $\theta$ simultaneously. The second model is $r(\rvx, \rvtheta_{\text{de}}) = {\pd(\rvx)}/{p_g(\rvx;\rvtheta_{\text{de}})}$, where $\rvtheta_{\text{de}}$ means we detach the gradient of $\rvtheta$ such that the second model cannot reflect the variability of $p_g$, i.e., the dependence between the ratio model and $p_g$ is discarded \citep{metz2016unrolled, franceschi2022neural}. The third model is $r_{\text{GAN}}(\rvx)$ where the density ratio is obtained by performing a single gradient update for the binary classification in standard GAN training. Note that $r(\rvx, \rvtheta_{\text{de}})$ and $r_{\text{GAN}}(\rvx)$ are only differentiable with $\rvx$.

We train the generator to minimize the following loss function with the above three density ratio models respectively (for $r_{\text{GAN}}(\rvx)$, we use the standard bi-level optimization),
\begin{equation} 
      \min_{\rvtheta} \mathbb{E}_{\rvz\sim p_\rvz} \big[ f(r)\big]\footnote{The Monte Carlo objective is resampled from $p_\rvz$ via the reparameterization trick \citep{kingma2013auto,rezende2014stochastic}.} \text{ or equivalently }\min_{\rvtheta} -\mathbb{E}_{\rvz\sim p_\rvz} \big[ h(\log r)\big]
\end{equation}
Given $f(r)$ we can rewrite it as a function of log density ratio $h(\log r)=-f(r)$. 
In this experiment, we consider five types of $f$-divergences with $f''(\cdot)>0$ (expressions summarized in Appendix \ref{b1}). In addition, we study a strictly increasing function with $h'(\cdot)>0$ given by $h(\log r) = \exp\big( 1.5 \log r \big)=r^{1.5}$ where its $f(r) = -r^{1.5}$ is concave.
The results are summarized in Table \ref{table_1}, which are consistent with our analysis that VDM is a convex problem requiring the ratio model $r(\rvx, \rvtheta)$ to have a dependence on the generator that allows for the functional optimization, whereas $r(\rvx, \rvtheta_{\text{de}})$ and $r_{\text{GAN}}(\rvx)$ works with increasing functions with $h'(\cdot)>0$ following the framework of MonoFlow. 

The difference between $r(\rvx, \rvtheta)$ and $r(\rvx, \rvtheta_{\text{de}})$ is that they result in different gradient estimations. Using the ratio model $r(\rvx, \rvtheta)$, the gradient to the parameter $\theta$ is evaluated by
 \begin{equation}
 \begin{split}
 \text{grad}(\theta) &= \mathbb{E}_{\rvz \sim p_{\rvz}}  \big[\nabla_{\theta} f(r(\rvx_\theta(\rvz), \theta))\big]\\ &=  \mathbb{E}_{\rvz \sim p_{\rvz}} \left[f'(r(\rvx_\theta(\rvz), \theta))\left(\frac{\partial r}{\partial \rvx_\theta} \frac{\partial \rvx_\theta}{ \partial\theta} + \frac{\partial r}{\partial \theta}\right)\right],
 \end{split}
 \end{equation}
 where backpropagation is applied to both $\rvx_\theta(\rvz)$ and $\theta$. However, to obtain the gradient estimation for the ratio model $r(\rvx, \rvtheta_{\text{de}})$ or $r_{\text{GAN}}(\rvx)$, backpropagation is only applied to the reparameterized sample $\rvx_\theta(\rvz)$,
\begin{equation}
\begin{split}
  \text{grad}(\theta) &=\mathbb{E}_{\rvz \sim p_{\rvz}}  \big[\nabla_{\theta} f(r(\rvx_\theta(\rvz), \theta_{\text{de}}))\big] \\ &=  \mathbb{E}_{\rvz \sim p_{\rvz}} \left[f'(r(\rvx_\theta(\rvz), \theta_{\text{de}}))\left(\frac{\partial r}{\partial \rvx_\theta} \frac{\partial \rvx_\theta}{ \partial\theta} \right)\right].
 \end{split}\label{31}
 \end{equation}
Eq.~(\ref{31}) is also compatible with $r_{\text{GAN}}(\rvx)$.

\textbf{Remark}: $r(\rvx, \rvtheta_{\text{de}})$ can recover the true $f$-divergence, but minimizing this $f$-divergence has no effects except for KL divergence. \citet{roeder2017sticking} showed that the obtained gradient estimation under KL divergence is still unbiased if detaching gradient operator is applied.

%% file: 5.Generator_loss.tex
\section{Algorithimic Insights: Alternatives of Generator Loss}
\subsection{Effectiveness of Generator Losses via Vector Field Rescaling}
In this part, we analyze the practical effectiveness of different types of generator losses. We provide a study for the discriminator trained under the binary classification problem since it outputs the log density ratio $d(\rvx)=\log r(\rvx)$ 
(see Table \ref{GAN_variants}). We consider five generator losses which are monotonically increasing functions of the log density ratio: \textbf{1). Vanilla loss}: $ h(d) = -\log(1 - \sigma(d))$; \textbf{2). Non-saturated (NS) loss}: $ h(d) = \log(\sigma(d))$. \textbf{3). Maximum likelihood estimation (MLE)}: $h(d) = \exp(d) $. \textbf{4). Logit loss}: $h(d)=d$. \textbf{5). Arcsinh loss}: $h(d)=\arcsinh(d)$ 

\begin{figure}[h]
     \centering
     \vspace{0mm}
     \label{generator_loss}
     \includegraphics[width=0.9\linewidth]{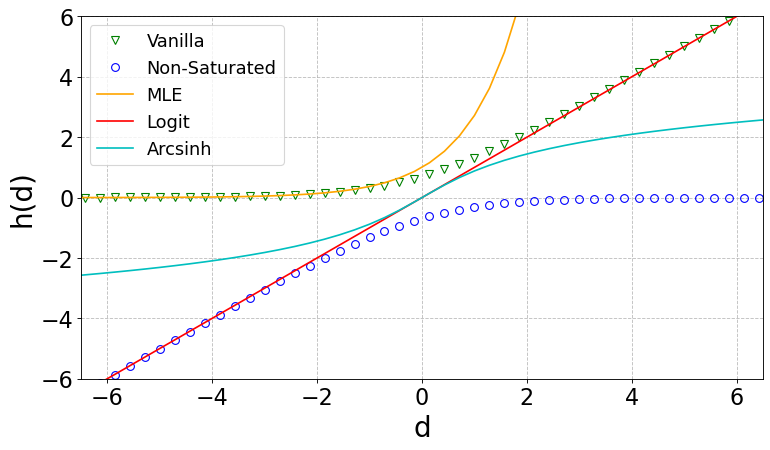}
     \vspace{-4mm}
     \caption{The plot of different generator losses as a function of $d$.}
     \label{gloss}
     \vspace{-3mm}
\end{figure}

The plot of these functions is shown in Figure \ref{gloss}. It is known that the vanilla loss and the MLE loss suffer from the gradient vanishing problem in practice \citep{goodfellow2016nips}. At the initial training steps, the generator is weak which means the associated $p_g$ is far away from $\pd$. If the discriminator is too good, the estimated log ratio can be extremely small, i.e., $d(\rvx) = \log \big[{\pd(\rvx)}/{p_g(\rvx)}\big] \ll 0$, for $\rvx \sim p_g(\rvx)$. We may observe in Figure \ref{gloss}, the curves of the vanilla loss and the MLE loss are fairly flat when $d(\rvx)\ll0$, which means the derivative $h'(\cdot)$ is nearly zero. According to Eq.~(\ref{mono gf}), such a rescaling scheme yields extremely small vector fields, resulting in the generator being trapped at the initial steps as the infinitesimal change of particles $\mathrm{d}\rvx_t \approx 0$. However, we may observe that the derivative of the vanilla loss and the ML loss deviates from zero if $d(x)$ is near zero. This suggests that these losses can work if the initial $p_g$ is close to $\pd$ where the estimated log ratio $d(\rvx) = \log \big[{\pd(\rvx)}/{p_g(\rvx)}\big]$ is not 
so small.

The NS loss, the logit loss and the arcsinh loss avoid gradient vanishing simply because they have non nearly zero derivatives when $d(\rvx)<0$ despite that the NS loss is flat when $d(\rvx)>0$. 
Since $\pd(\rvx)/p_g(\rvx) \approx 0$ for $\rvx \sim p_g(\rvx)$ at the beginning,  
the log ratio $d(\rvx)$ gradually increases from a negative value to zero during the training. When $d(\rvx)$ approaches zero, it means $p_g\approx\pd$ such that the generator has learned the data distribution.

\subsection{An Embarrassingly Simple Trick to Fix Vanilla GAN on MNIST Generation}
\begin{figure}[h]
     \centering
     \includegraphics[width=0.8\linewidth]{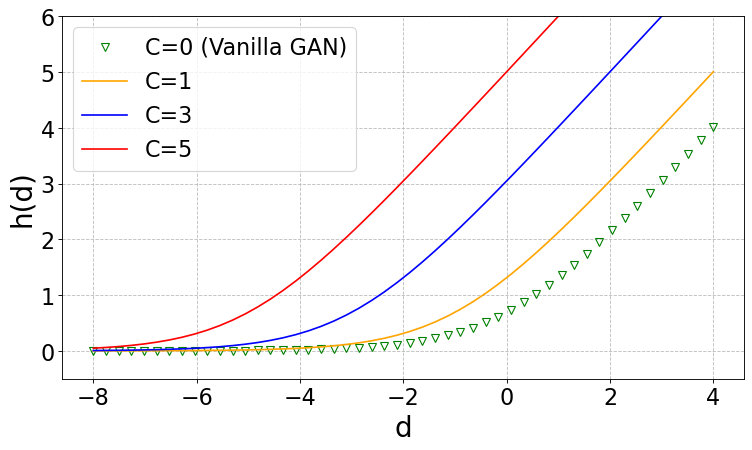}
     \vspace{-4mm}
     \caption{The plot of the vanilla losses by adding different $C$s.}
     \vspace{-2mm}
     \label{add_constant}
\end{figure}

\begin{figure}[!htb]
\vspace{-0.2cm}
     \centering
     \begin{subfigure}{0.16\textwidth}
         \centering
         \includegraphics[width=\textwidth]{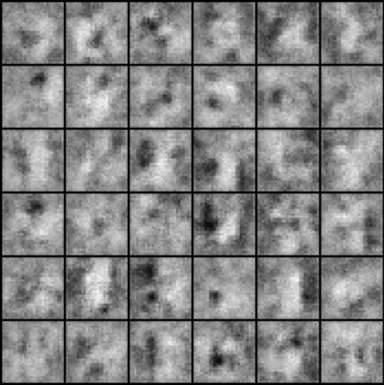}
         \vspace{-4mm}
         \caption{$C=0$}
     \end{subfigure}
     \begin{subfigure}{0.16\textwidth}
         \centering
         \includegraphics[width=\textwidth]{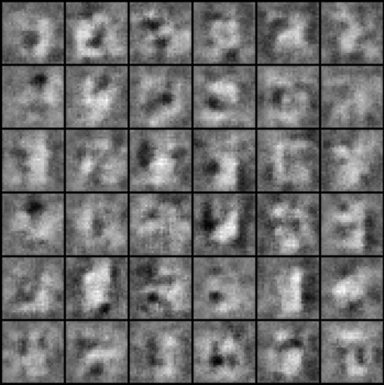}
         \vspace{-4mm}
         \caption{$C=1$}
         
     \end{subfigure}
     \begin{subfigure}{0.16\textwidth}
         \centering
         \includegraphics[width=\textwidth]{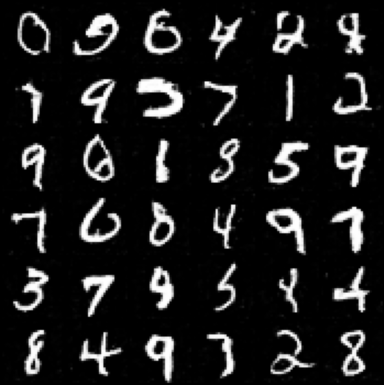}
         \vspace{-4mm}
         \caption{$C=3$}
     \end{subfigure}
     \begin{subfigure}{0.16\textwidth}
         \centering
         \includegraphics[width=\textwidth]{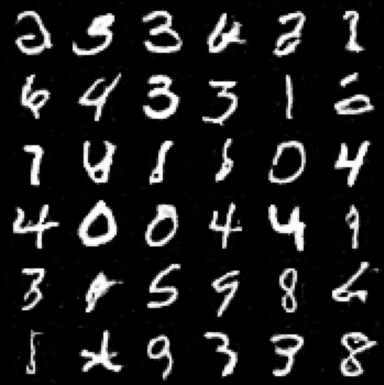}
         \caption{$C=5$}
     \end{subfigure}
     \vspace{-2mm}
             \caption{Generated samples with different $C$s. }
        \label{images}

\end{figure}

We have justified that MonoFlow can work with any strictly increasing mappings of the log density ratio and this mapping's derivative should deviate from zero when the $d(\rvx) < 0$ to better avoid too small rescaled vector fields (gradient vanishing). We show the effects of shifting the generator loss of the vanilla GAN left by adding a constant $C$ to the Sigmoid function,
\begin{equation}
         h(d) = -\log(1 - \sigma(d+C)) \label{generator loss}
\end{equation}
By adding a constant, we can obtain an increasing function whose derivative deviates from zero significantly, see Figure \ref{add_constant}. The neural network architecture used here is DCGAN \citep{radford2015unsupervised} and we follow the vanilla GAN framework where the log density ratio is obtained by logit output from the binary classifier and the model is trained with 15 epochs. The generated samples are shown in Figure \ref{images}. We observe that when $C=3$ and $C=5$, the generator losses in Eq.~(\ref{generator loss}) start to work, i.e., generators output plausible fake images.

\vspace{-0.1cm}

%% file: 6.RW_con.tex
\section{Related Works}
\textbf{Gradient Flow}: 
Wasserstein gradient flows of $f$-divergences have been previously studied in deep generative modeling as a refinement approach to improve sample quality \citep{ansari2020refining}. A close work to ours is \citep{gao2019deep} where the authors proposed to use gradient flows of $f$-divergences to refine fake samples output by the generator and the generator learns to minimize the squared distance between the refined samples and the original fake samples. However, neither of the above reveals the equivalence between gradient flows and divergence GANs. Furthermore, MonoFlow is a more generalized framework to cover existing gradient flows of $f$-divergences and our method also applies to traditional loss designs as well as many other types of monotonically increasing functions.
\textbf{IPM GANs}: Our framework unifies divergence GANs since estimating a probability divergence is naturally related to density ratio estimation \citep{sugiyama2012density}. However, some variants of GANs are developed with Integral Probability Metric (IPM) \citep{sriperumbudur2009integral}. For example, WGANs \citep{arjovsky2017wasserstein, gulrajani2017improved} estimate the Wasserstein-1 metric and then minimize this metric. While MonoFlow is associated with Wasserstein-2 metric, minimizing a functional in $\mathcal{P}(\mathbb{R}^{d})$ naturally decreases Wasserstein-2 metric as well. Other types of IPM GANs are MMD GAN \citep{dziugaite2015training} and Sobolev GAN \citep{mroueh2017sobolev}. Both of them have been interpreted as gradient flow approaches \citep{mroueh2021convergence, mroueh2018sobolev} but associated with different vector fields. \citet{franceschi2022neural} studied the NTK view on GANs given a vector field specified by a loss function of IPM but lacks connections to divergence GANs. 
\textbf{Diffusion Models}: diffusion models \citep{ho2020denoising, song2020score, luo2022understanding} are another line of generative modeling framework. This framework first perturbs data by adding noises with different scales to create a path $\{q_t\}_{t\geq 0}$ interpolating the data distribution and the noise distribution. Subsequently, the generative modeling is to reverse $\{q_t\}_{t\geq 0}$ as denoising. The similarity between MonoFlow and diffusion models is that they both involve particle evolution associated with different paths of marginal probabilities. However, the vector field of MonoFlow is obtained with the log density ratio that must be corrected per iteration by gradient update, whereas diffusion models directly estimate vector fields by time-dependent neural networks and they are straightforward particle methods.

\section{Conclusions}
MonoFlow provides a unified framework to explain why and how adversarial training of divergence GANs works. The mechanism of adversarial training may not be as adversarial as we used to think. It instead simulates an ODE system. The bi-level step of adversarial can be regarded as first estimating the vector field, then updating the generator as learning to draw particles of the ODE, a process we call parameterizing MonoFlow. All divergence GANs discussed in this paper are unified  under our framework. They all are different methods of estimating the bijection of the log density ratio and then mapping the log density ratio by different monotonically increasing functions. ,
The methodological development closely matches our theoretical framework. The limitation of this paper is that our framework does not cover IPM GANs since these variants give a vector field that is different from the gradient of log density ratios. We leave it as a future work.

%% file: 7.Appendix.tex
\section{Appendix}\label{appA}
\subsection{Proof of Theorem \ref{theorem1}}
\label{proof.theorem1}



The dissipation rate:
For any curve $\{q_t\}_{t \geq 0}$ evolving according to the vector field $\{v_t\}_{t\geq 0}$, the dissipation rate of the functional \citep{ambrosio2005gradient} is given as 
\begin{equation}
    \frac{\partial\mathcal{F}(q_t)}{\partial t} = \int {\langle \nabla_{W_2} {\mathcal{F}(q_t)}, v_t \rangle} \mathrm{d}q_t.
\end{equation}
If the functional is the KL divergence, we have the associated Wasserstein gradient  $\nabla_{W_2} {\mathcal{F}(q_t)} = \gradx \log (q_t/p)$. Recall that the vector filed of MonoFlow in Eq.~(\ref{mono gf}) is 
\begin{equation}
    v_t = h'(\log r_t)\nabla_{\rvx} \log r_t, \quad r_t = \log (p/q_t)
\end{equation}

Therefore, if $h'(\cdot) > 0$, MonoFlow dissipates the KL divergence with the rate
\begin{equation}
     \frac{\partial\mathcal{F}(q_t)}{\partial t} = \mathbb{E}_{\rvx \sim q_t}\left[-h'\big(\log r_t (\rvx) \big){\left\lVert\gradx \log  \frac{q_t(\rvx)}{p(\rvx)} \right\rVert^2}\right] \leq 0.
\end{equation}
It is obvious that  ${\partial\mathcal{F}(q_t)}/{\partial t} =0 $ implies 
\begin{equation}
    \mathbb{E}_{\rvx \sim q_t}\left[{\big\lVert\gradx \log  q_t(\rvx) - \gradx\log  p(\rvx)\big\rVert^2}\right] = 0,
\end{equation}
where the left hand side is the Fisher divergence $\mathcal{D}_{\rm{FI}}(q_t||p)$. If $p$ is a well defined proper probability measure on $\mathbb{R}^n$, $\mathcal{D}_{\rm{FI}}(q_t||p)$ attains zero if and only if $q_t=p$.

 Hence, MonoFlow always decreases the KL divergence with the time when $q_t\neq p$. By the monotone convergence theorem, i.e., any decreasing sequence converges to its infimum, thus the KL divergence converges to zero if $t\to \infty$, which indicates $q_t$ evolves to $p$.

\subsection{Proof of Theorem \ref{theorem2}}
\label{proof.theorem2}
Define the functional $\gF(q)$ of $f$-divergences as
\begin{equation}
\gF(q) = \mathcal{D}_f(p||q) = \int f\left(\frac{p}{q}\right)(\rvx) q(\rvx)\mathrm{d}\rvx.
\end{equation}
where $f\colon \mathbb{R}^+ \to \mathbb{R}$ is a convex function and we may further assume that $f$ is twice differentiable.

Let $\phi \in \mathcal{P}(\mathbb{R}^n) $ be a test function, the first variation (functional derivative) $\frac{\delta \gF}{\delta q}$ is defined as
\begin{equation}
\begin{split}
\int \frac{\delta \gF}{\delta q} (\rvx) \phi(\rvx) \mathrm{d}\rvx &= \lim_{\epsilon \to 0} \frac{\gF(q+\epsilon\phi)- \gF(q) }{\epsilon} \\
&=\frac{d}{d\epsilon} \gF(q+\epsilon\phi)\Big \vert_{\epsilon=0}\\
& = \frac{d}{d\epsilon} \int f\left(\frac{p}{q+\epsilon\phi}\right)(\rvx)\big(q(\rvx)+\epsilon\phi(\rvx)\big) \mathrm{d}\rvx\Big \vert_{\epsilon=0} \\
&=  \int  \left\{f\left(\frac{p}{q+\epsilon\phi}\right) (\rvx) \phi(\rvx)- f'\left(\frac{p}{q+\epsilon\phi}\right)(\rvx) \frac{p(\rvx) \phi(\rvx)}{q(\rvx)+\epsilon\phi(\rvx)} \right\} \mathrm{d}\rvx   \Big \vert_{\epsilon=0} \\
&=  \int  \left\{f\left(\frac{p}{q}\right) - f'\left(\frac{p}{q}\right) \frac{p}{q} \right\} (\rvx)\phi(\rvx) \mathrm{d}\rvx.  \\
\end{split}
\end{equation}
Thus,
\begin{equation}
\frac{\delta \gF}{\delta q} = f(r) - rf'(r), \quad \text{where} \quad r = \frac{p}{q} .\label{frf}
\end{equation}
Recall that the Wasserstein gradient of $\gF(q)  $ is the Euclidean gradient of the first variation, we have 
\begin{equation}
\nabla_{W_2} \gF(q) = \gradx \frac{\delta \gF}{\delta q} = - r f''(r) \gradx r.
\end{equation}
The corresponding vector field is given by the negative Euclidean gradient, see Section \ref{wgf}, therefore the particle flow ODE of $f$-divergences can be written as  
\begin{equation}
\mathrm{d}\rvx = -\gradx \frac{\delta \gF}{\delta q}(\rvx)\mathrm{d}t  = r(\rvx) f''(r(\rvx)) \gradx r(\rvx)\mathrm{d}t = r(\rvx)^2 f''(r(\rvx)) \gradx \log r(\rvx)\mathrm{d}t.
\end{equation}

\subsection{Proof of Corollary 3.3}
\label{proof.coro}
According to the existence theorem of primitive functions (antiderivative), any  continuous scalar function must have a primitive function and this primitive function is also continuous. Hence, if $f''(r) = h'(\log r) /r^2$ where $h$ is continuously differentiable, we have $f''(r)$ is continuous such that $f'(r)$ and $f(r)$ both exist. We also have $h'(\log r)>0\Longrightarrow f''(r)>0$, this indicates $f$ is strictly convex. 

Hence, given a differentiable $h$ with $h'(\cdot) > 0$, there must exist a strictly convex function $f$ such that $f(1)=0$ (primitive functions differ in constants) where its second derivative is specified by $f''(r) = h'(\log r) /r^2$.

We can let $h(\log r) = rf'(r) - f(r)+C$, apparently $h(\log r) = rf'(r) - f(r)+C \Longleftrightarrow h'(\log r) = r^2f''(r)$, this defines the particle evolution of Wasserstein gradient flows, 
\begin{equation}
    \mathrm{d}\rvx_t 
     = h'\big(\log r_t(\rvx_t)\big)\nabla_{\rvx} \log r_t(\rvx_t) \mathrm{d}t = r(\rvx_t)^2f''\big(r(\rvx_t)\big)\nabla_{\rvx} \log r_t(\rvx_t) \mathrm{d}t.
\end{equation}
 Without the loss of generality, we can let $h(\log r) = rf'(r) - f(r)$.

\subsection{Proof of Lemma 3.4} \label{assumption}

This proof is adapted from Lemma 1 and 2 of \citet{moustakides2019training}.
Given the optimization problem
\begin{equation}
\max_{d \in \mathcal{H}} \ \mathbb{E}_{\rvx\sim p}\left[\phi\big(d(\rvx)\big)\right] + \mathbb{E}_{\rvx\sim q}\left[\psi \big( d(\rvx)\big)\right],
\end{equation}
where $\mathcal{H}$ is a class of all measurable functions.
We rewrite it as 
\begin{equation}
\begin{split}
& \max_{d \in \mathcal{H}} \  \mathbb{E}_{\rvx\sim q}\left[\frac{p(\rvx)}{q(\rvx)} \phi\big(d(\rvx)\big) + \psi \big( d(\rvx)\big)\right] \\
&= \mathbb{E}_{\rvx\sim q}\left[\max_{d \in \mathcal{H}} \left\{\frac{p(\rvx)}{q(\rvx)} \phi\big(d(\rvx)\big) + \psi \big( d(\rvx)\big) \right\} \right], 
\end{split}
\end{equation}
we apply the interchange of maximum and integral because the integral operator is independent of $d$.  
Since the maximum is holding for every fixed $\rvx$, thus we let the derivative $\frac{\partial }{\partial d(\rvx)} \left[\frac{p(\rvx)}{q(\rvx)} \phi\big(d(\rvx)\big) + \psi \big( d(\rvx)\big)\right] = 0$, we have the optimal $d^\ast$, the abbreviation of $d^\ast(\rvx)$, to satisfy
\begin{equation}
r\phi'(d^\ast) + \psi' (d^\ast) = 0, \quad r(\rvx) = \frac{p(\rvx)}{q(\rvx)} >0 \label{ratioEq.}
\end{equation}

Furthermore, we need to discuss under what sufficient conditions, $d^\ast$ is the unique maximizer for the above problem. Denote $l(d) = r\phi(d) + \psi(d)$, in order to ensure that $d^\ast$ is the unique maximizer, $l'(d)$ should satisfy 
\begin{equation}
l'(d)>0, \forall d<d^\ast \text{ and } l'(d)<0, \forall d>d^\ast.  \label{condi}
\end{equation}
We define the mapping $\mathcal{T}(d) : = -\frac{\psi'(d)}{\phi' ( d) }$ and summarize two \textbf{sufficient conditions} as:
\begin{enumerate}
    \item $\phi$ is concave, $\psi$ is strictly concave and the resulting mapping $\mathcal{T}$ is a bijection.
    \item $\phi'(\cdot) >0 $ and the resulting mapping $\mathcal{T}$ is a strictly increasing mapping (also a bijection).
\end{enumerate}
It is obvious that if $\mathcal{T}$ is a bijection, $d^\ast=\mathcal{T}^{-1}(r)$ is the root of Eq.~(\ref{ratioEq.}).

For condition 1, since $\phi$ is concave and $\psi$ is strictly concave, the linear combination $l(d)$ is strictly concave which satisfies Eq.~(\ref{condi}). Therefore, $d^\ast$ is the unique maximizer.

For condition 2, we can write $l'(d) = [\mathcal{T} (d^*) - \mathcal{T} (d) ]\phi'(d)$. Since $\mathcal{T}$ is a strictly increasing mapping and $d^*$ is the maximizer, we have $\mathcal{T} (d^*) - \mathcal{T} (d)>0$ for $d<d^\ast$ and $\mathcal{T} (d^*) - \mathcal{T} (d)<0$ for $d>d^\ast$. Hence $l'(d)$ satisfies the condition stated in Eq.~(\ref{condi}).

In Table \ref{table_1}, $b$-gan satisfies condition 2 and the rest of the divergence GANs satisfy condition 1.

Some examples:
\begin{itemize}
    \item For binary classification, $\phi(d) = \log \sigmoid(d)$ and $\psi(d) = \log (1-\sigmoid(d))$, $r(\rvx)=\exp(d^*(\rvx))$.
    \item Fenchel-duality, $\phi(d) = d $, $\psi(d)=-\tilde{f}(d) $, $r(\rvx) = {\tilde{f}}' \big(d^*(\rvx)\big)$ where the convex conjugate is $\tilde{f}(d) = \sup_{r\in\text{dom}f}\{rd - f(r)\}$
    \item For least-square GAN, $\phi(d) = -(d -1) ^2, \psi(d)=-d^2$, $r(\rvx) = \frac{d^*(\rvx)}{1-d^*(\rvx)}$
\end{itemize}

\section{Variational Divergence Minimization}
\label{videtail}
Given an $f$-divergence $\mathcal{D}_f(p||q)$ where $p$ is the fixed target distribution, variational divergence minimization finds an approximating distribution $q$ via the functional optimization
\begin{equation}
\min_q \mathcal{D}_f(p||q).
\end{equation}
If $q$ is represented by a parametric model with an explicit density function $q(\rvx;\theta)$, the $f$-divergence can be written as 
\begin{equation}
\text{Cost}(\theta) = \mathcal{D}_f(p||q) =  \int f\big(  r(\rvx, \theta) \big) q(\rvx;\theta)\mathrm{d}\rvx
\end{equation}
where $r(\rvx, \theta) = {p(\rvx)}/{q(\rvx;\theta)}$ explicitly depends on $\rvx$ and $\theta$. The $f$-divergence becomes a cost function of the parameter $\theta$ because $\rvx$ is integrated out. A typical example is in standard variational inference where $q$ is a parametric Gaussian distribution such that we know the exact density function $q(\rvx;\theta)$. 

Hence, the functional optimization problem degenerates into an optimization problem over the parameter space, 
\begin{equation}
\min_q \mathcal{D}_f(p||q) \Longleftrightarrow \min_{\theta} \text{Cost}(\theta). \label{vdm_cost}
\end{equation}

Since $f$ is convex, we can apply Jensen's inequality, 
\begin{equation}
\mathcal{D}_f(p||q) = \int f\left( \frac{p(\rvx)}{q(\rvx;\theta)} \right) q(\rvx;\theta)\mathrm{d}\rvx  \geq f\left(\int \frac{p(\rvx)}{q(\rvx;\theta)} q(\rvx;\theta) \mathrm{d}\rvx \right) =f(1)=0,
\end{equation}
Jensen's inequality indicates that $\mathcal{D}_f(p||q)=0$ if and only if ${p(\rvx)}/{q(\rvx;\theta)}$ is a constant, such that we have $q=p$.\\\\
Therefore, we can minimize the cost function $\text{Cost}(\theta)$ to approximate $p$ with $q$. Solving the optimization problem in Eq.~(\ref{vdm_cost}) requires Monte Carlo gradient estimation, we can apply the reparameterization trick \citep{kingma2013auto, rezende2014stochastic} to solve 
\begin{equation}
    \min_\theta \text{Cost}(\theta)= \min_\theta \mathbb{E}_{\rvz \sim p_{\rvz}} \big[f\big(r(\rvx_\theta(\rvz), \theta)\big)\big], \text{ where } \rvx_\theta(\rvz) = g_\theta(\rvz).
\end{equation}
 $g_\theta$ is the Gaussian generator parameterized by $\theta$. The associated gradient estimation is obtained by the chain rule, 
 \begin{equation}
 \mathbb{E}_{\rvz \sim p_{\rvz}}  \big[\nabla_{\theta} f(r(\rvx_\theta(\rvz), \theta))\big] =  \mathbb{E}_{\rvz \sim p_{\rvz}} \big[f'(r(\rvx_\theta(\rvz), \theta))(\frac{\partial r}{\partial \rvx_\theta} \frac{\partial \rvx_\theta}{ \partial\theta} + \frac{\partial r}{\partial \theta})\big]
 \end{equation}
 if we detach the gradient of $\theta$ in the ratio model, the gradient estimation is distorted,
  \begin{equation}
 \mathbb{E}_{\rvz \sim p_{\rvz}}  \big[\nabla_{\theta} f(r(\rvx_\theta(\rvz), \theta_{\text{de}}))\big] =  \mathbb{E}_{\rvz \sim p_{\rvz}} \big[f'(r(\rvx_\theta(\rvz), \theta_{\text{de}}))(\frac{\partial r}{\partial \rvx_\theta} \frac{\partial \rvx_\theta}{ \partial\theta} )\big].
 \end{equation}

 Alternatively, we can apply score function gradient estimation, see \citep{mohamed2019monte} for more details.
\newpage
\section{Experiments}
All codes are available at \url{https://github.com/YiMX/MonoFlow}.
\subsection{Experiment Details for Section \ref{4.3}} \label{b1}
\begin{figure*}[!htb]
\vspace{-0.0cm}
     \centering
     \begin{subfigure}{0.34\textwidth}
         \centering
         \includegraphics[width=\textwidth]{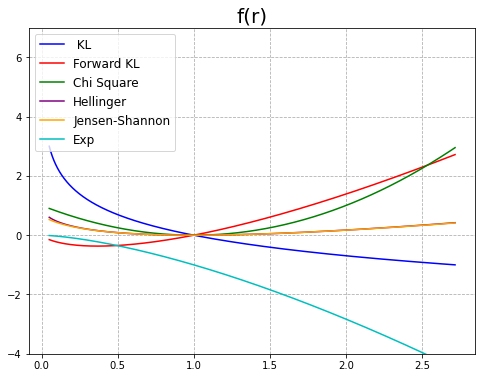}
         \vspace{-4mm}
     \end{subfigure}
     \begin{subfigure}{0.34\textwidth}
         \centering
         \includegraphics[width=\textwidth]{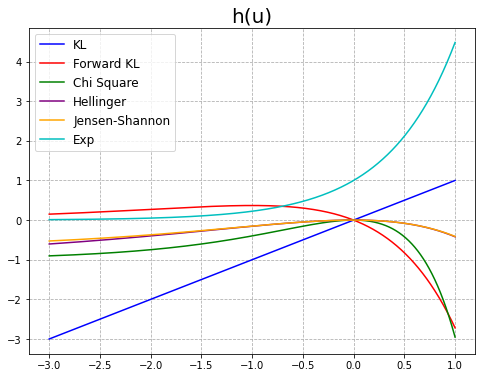}
         \vspace{-4mm}
     \end{subfigure}
     \caption{Function plots of $f(r)$ and $h(u)$}
     \label{funcs}
\end{figure*}

\begin{table}[h]
\caption{Explicit forms of $f$ and $h$}
\label{expilcit form}
\begin{center}
\begin{tabular}{llll}
\multicolumn{1}{c}{}  &\multicolumn{1}{c}{$f(r)$}  & $h(u), u=\log r$ \\
\hline \\
KL             &$-\log r$                                     & $u$\\
Forward KL             &$r\log r$                                     & $-u \exp(u)$ \\
Chi-Square             &$(r-1)^2$                                     & $-(\exp(u) -1)^2$\\
Hellinger              &$(\sqrt{r}-1)^2$                              & $-(\sqrt{\exp(u)} -1)^2$\\
Jensen-Shannon (GAN)   &$r \log \frac{2r}{1+r} + \log \frac{2}{1+r}$  &$ -\exp(u) \log \frac{2\exp(u)}{1+\exp(u)} - \log \frac{2}{1+\exp(u)}$\\
Exp                       &  $-\exp(1.5 \log r)$          & $\exp(1.5 u)$  \\

\end{tabular}
\end{center}
\end{table}
In Figure \ref{funcs}, we can observe that the Exp function is concave under $f(r)$. KL and Exp are increasing functions under $h(u)$.
 \begin{align*}
 \text{The generator is initialized at: } 
 N
\begin{bmatrix}
\begin{pmatrix}
1.0\\
1.0
\end{pmatrix},
\begin{pmatrix}
1.00 & 0.00 \\
0.00 & 1.00 
\end{pmatrix}
\end{bmatrix}
\text{and the target distribution is : } 
 N
\begin{bmatrix}
\begin{pmatrix}
0.0\\
0.0
\end{pmatrix},
\begin{pmatrix}
1.00 & 0.80 \\
0.80 & 0.89 
\end{pmatrix}
\end{bmatrix}
\end{align*}
$r_{\text{GAN}}(\rvx)$ uses a simple 2-layer discriminator with Leaky ReLU activation that has logit output as the log density ratio. 
\newpage
\subsection{Visualization Results for Section \ref{4.3}}  \label{visu}
\vspace{-2mm}
\begin{figure}[htb!]
     \makebox[20pt]{\raisebox{40pt}{\rotatebox[origin=c]{90}{KL}}}%
     \begin{subfigure}[b]{0.31\textwidth}
         \centering
         \caption{Ratio model: $r(\rvx, \theta)$}
         \includegraphics[width=\textwidth]{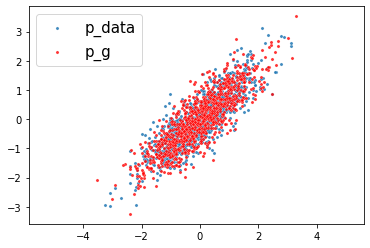}
     \end{subfigure}
     \begin{subfigure}[b]{0.31\textwidth}
         \centering
         \caption{Ratio model: $r(\rvx, \theta_{\text{de}})$}
         \includegraphics[width=\textwidth]{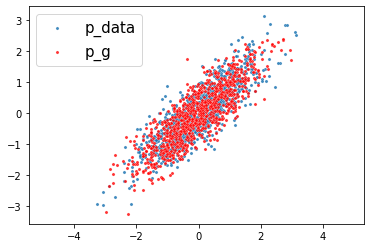}
     \end{subfigure}
     \begin{subfigure}[b]{0.31\textwidth}
         \centering
          \caption{Ratio model: $r_{\text{GAN}}(\rvx)$}
         \includegraphics[width=\textwidth]{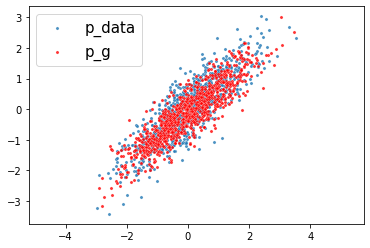}
     \end{subfigure}
     \makebox[20pt]{\raisebox{40pt}{\rotatebox[origin=c]{90}{Forward-KL}}}%
\begin{subfigure}[b]{0.31\textwidth}
         \centering
         \includegraphics[width=\textwidth]{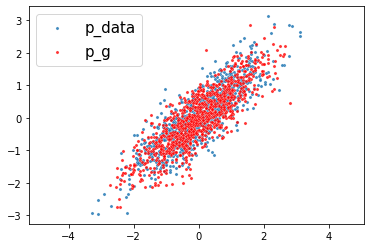}
     \end{subfigure}
     \begin{subfigure}[b]{0.31\textwidth}
         \centering
         \includegraphics[width=\textwidth]{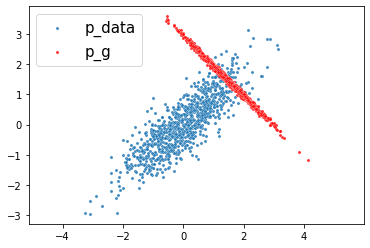}
     \end{subfigure}
     \begin{subfigure}[b]{0.31\textwidth}
         \centering
         \includegraphics[width=\textwidth]{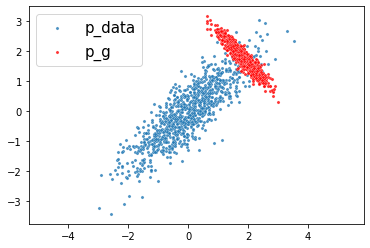}         
     \end{subfigure}
     \makebox[20pt]{\raisebox{40pt}{\rotatebox[origin=c]{90}{Chi-Square}}}%
\begin{subfigure}[b]{0.31\textwidth}
         \centering
         \includegraphics[width=\textwidth]{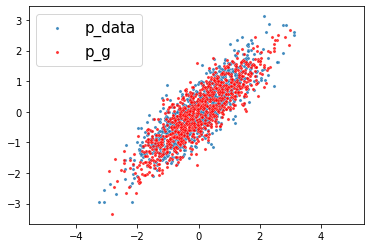}
     \end{subfigure}
     \begin{subfigure}[b]{0.31\textwidth}
         \centering
         \includegraphics[width=\textwidth]{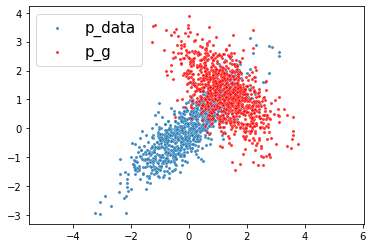}
     \end{subfigure}
     \begin{subfigure}[b]{0.31\textwidth}
         \centering
         \includegraphics[width=\textwidth]{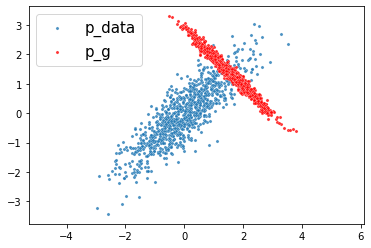}
     \end{subfigure}
     \makebox[20pt]{\raisebox{40pt}{\rotatebox[origin=c]{90}{Hellinger}}}%
  \begin{subfigure}[b]{0.31\textwidth}
         \centering
         \includegraphics[width=\textwidth]{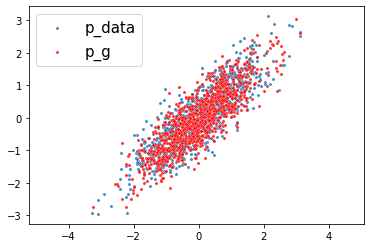}
     \end{subfigure}
     \begin{subfigure}[b]{0.31\textwidth}
         \centering
         \includegraphics[width=\textwidth]{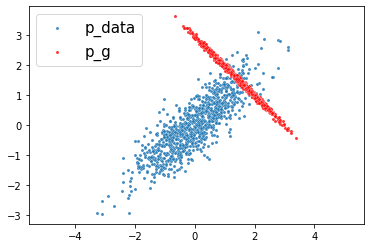}
     \end{subfigure}
     \begin{subfigure}[b]{0.31\textwidth}
         \centering
         \includegraphics[width=\textwidth]{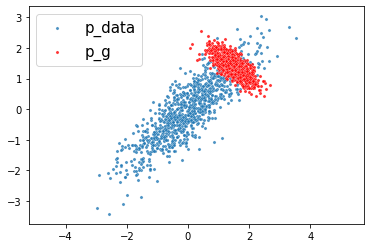}
     \end{subfigure}
     \makebox[20pt]{\raisebox{40pt}{\rotatebox[origin=c]{90}{Jensen-Shannon}}}%
     \begin{subfigure}[b]{0.31\textwidth}
         \centering
         \includegraphics[width=\textwidth]{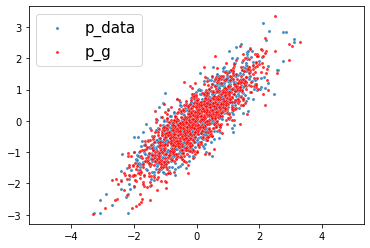}
     \end{subfigure}
     \begin{subfigure}[b]{0.31\textwidth}
         \centering
         \includegraphics[width=\textwidth]{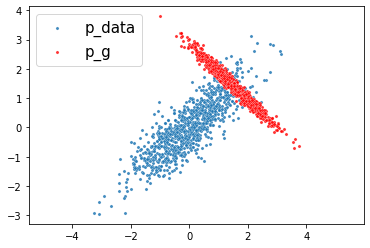}
     \end{subfigure}
     \begin{subfigure}[b]{0.31\textwidth}
         \centering
         \includegraphics[width=\textwidth]{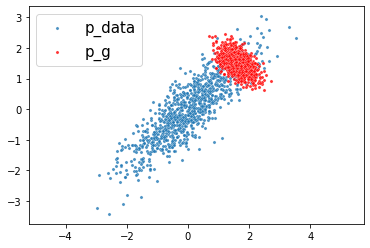}
     \end{subfigure}
    \makebox[20pt]{\raisebox{40pt}{\rotatebox[origin=c]{90}{Exp}}}%
     \begin{subfigure}[b]{0.31\textwidth}
         \centering
         \includegraphics[width=\textwidth]{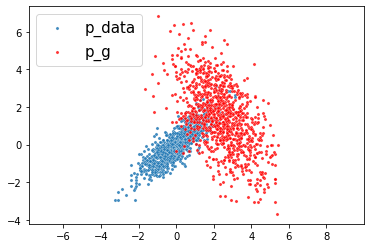}
     \end{subfigure}
     \hspace{0.9mm}
     \begin{subfigure}[b]{0.31\textwidth}
         \centering
         \includegraphics[width=\textwidth]{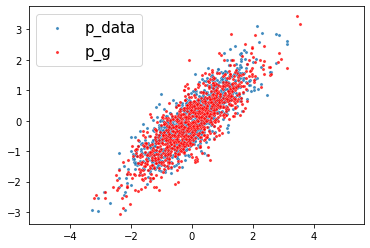}
     \end{subfigure}
     \hspace{0.9mm}
     \begin{subfigure}[b]{0.31\textwidth}
         \centering
         \includegraphics[width=\textwidth]{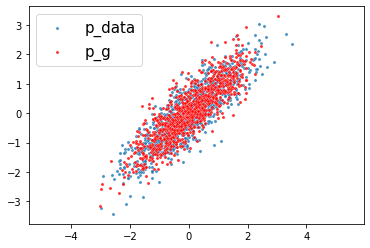}
     \end{subfigure}
     
\end{figure}
\newpage
        
         
         
\newpage
 
\subsection{FID Scores of Different Generator Losses}
In this section, we demonstrate that different generator losses can achieve equal performances on image generations by evaluating the FID scores \citep{heusel2017gans}. 
The discriminator are trained with the original GAN objective \citep{goodfellow2014generative} where the optimal $d^*(\rvx) = \log r(\rvx)$ and the Least-square GAN \citep{mao2017least} objective where the optimal $d^*(\rvx)={r(\rvx)}/({1+r(\rvx)})$, see Table \ref{table_1}. We use MNIST,  CIFAR-10 \citep{krizhevsky2009learning} and  Celeb-A \citep{liu2015deep} datasets in this experiment. Models are trained using the training sets and the FID scores are evaluated on the test sets as shown in Table \ref{t4} and Table \ref{t5}. The neural network structures are modified from \citep{radford2015unsupervised}. 

\begin{table}[h]

\caption{FID scores with different generator losses where $d^*(\rvx)=\log r(\rvx)$}
\vspace{-0mm}
\begin{center}
\begin{tabular}{rccc}
\label{t4}
$h(d(\rvx))$	&MNIST	&CIFAR-10	&Celeb-A\\
\toprule
$\log \sigma (d(\rvx))$: &4.3309	&21.2980	&20.6109\\
$d(\rvx)$ : &4.4631	&20.7969	&21.2240\\
$\arcsin (d(\rvx))$: &4.4893	&21.2533	&21.0077\\
\bottomrule
\end{tabular}
\end{center}
\vspace{-3mm}
\end{table}

\begin{table}[h]

\caption{FID scores with different generator losses where $d^*(\rvx)=\frac{r(\rvx)}{1+r(\rvx)}$}
\vspace{-0mm}
\begin{center}
\begin{tabular}{rccc}
\label{t5}
$h(d(\rvx))$	&MNIST	&CIFAR-10	&Celeb-A\\
\toprule
$-(d(\rvx)-1)^2$: &5.0808	&23.8330	&21.6787\\
$d(\rvx)$: &4.6000	&22.7969	&20.5231\\
$\arcsin (d(\rvx))$:&4.5525	&23.4698	&22.1024\\
\bottomrule
\end{tabular}
\end{center}
\vspace{-3mm}
\end{table}